\documentclass[11pt,a4paper,twocolumn]{article}
\usepackage[utf8]{inputenc}
\usepackage{amsmath}
\usepackage{amssymb}
\usepackage{hyperref}
\usepackage{natbib}
\usepackage{geometry}
\usepackage{booktabs}
\usepackage{graphicx}
\usepackage{enumitem}
\title{\bfseries Multi-LLM Thematic Analysis with Dual Reliability Metrics: Combining Cohen's Kappa and Semantic Similarity for Qualitative Research Validation}

\author{
Nilesh Jain$^{1,2}$ \and
Hyungil Suh$^3$ \and
Seyi Adeyinka$^1$ \and
Leor Roseman$^3$ \and
Aza Allsop$^{1,2}$ \\[0.3em]
\footnotesize $^1$AZA Lab, Department of Psychiatry, Yale University, \\
\footnotesize \hspace{0.5em} New Haven, CT, United States \\
\footnotesize $^2$Department of Psychiatry and Behavioral Sciences, \\
\footnotesize \hspace{0.5em} Center for Collective Healing, Howard University, \\
\footnotesize \hspace{0.5em} Washington, DC, United States \\
\footnotesize $^3$Department of Psychology, University of Exeter, \\
\footnotesize \hspace{0.5em} Exeter, United Kingdom
}

\date{\today}

\begin{document}

\twocolumn[
\begin{@twocolumnfalse}
\maketitle

\begin{abstract}
Qualitative research faces a critical reliability challenge: traditional inter-rater agreement methods require multiple human coders, are time-intensive, and often yield moderate consistency. We present a multi-perspective validation framework for LLM-based thematic analysis that combines ensemble validation with dual reliability metrics: Cohen's Kappa ($\kappa$) for inter-rater agreement and cosine similarity for semantic consistency. Our framework enables configurable analysis parameters (1-6 seeds, temperature 0.0-2.0), supports custom prompt structures with variable substitution, and provides consensus theme extraction across any JSON format. As proof-of-concept, we evaluate three leading LLMs (Gemini 2.5 Pro, GPT-4o, Claude 3.5 Sonnet) on a psychedelic art therapy interview transcript, conducting six independent runs per model. Results demonstrate Gemini achieves highest reliability ($\kappa = 0.907$, cosine=95.3\%), followed by GPT-4o ($\kappa = 0.853$, cosine=92.6\%) and Claude ($\kappa = 0.842$, cosine=92.1\%). All three models achieve a high agreement ($\kappa > 0.80$), validating the multi-run ensemble approach. The framework successfully extracts consensus themes across runs, with Gemini identifying 6 consensus themes (50-83\% consistency), GPT-4o identifying 5 themes, and Claude 4 themes. Our open-source implementation provides researchers with transparent reliability metrics, flexible configuration, and structure-agnostic consensus extraction, establishing methodological foundations for reliable AI-assisted qualitative research.
\end{abstract}

\vspace{0.5cm}
\end{@twocolumnfalse}
]

\section{Introduction}

Inter-rater reliability remains a fundamental challenge in qualitative research \citep{braun2006using}. Traditional approaches require multiple human coders who independently analyze the same data, with agreement measured through Cohen's kappa. This process is time-intensive, expensive, and often yields only moderate agreement ($\kappa = 0.40$-$0.60$). The emergence of large language models (LLMs) offers potential solutions, but current approaches exhibit several limitations.

Recent LLM-based systems have evolved from simple prompting to complex architectures. Frameworks like QualIT \citep{kapoor2024qualit} utilize iterative clustering, while multi-agent systems like Thematic-LM \citep{qiao2025thematic} and TAMA \citep{xu2025tama} leverage agentic interactions to enhance the \textit{quality} of thematic extraction.

However, a critical gap remains in reliability quantification. While these methods employ multiple inference steps to refine outputs, they typically present the final result as a definitive conclusion without assessing its stochastic stability. They rarely quantify how sensitive the analysis is to the model's inherent probabilistic nature. Without measuring variance across independent runs, it remains unclear whether a generated theme is a robust finding or an artifact of a specific interaction path. 

As noted in reasoning tasks \citep{wang2022self}, single-point generation fails to capture the model's full interpretive distribution. Consequently, single-run analyses lack statistical robustness and fail to distinguish between stable thematic signals and hallucinatory noise. 

We propose a multi-perspective validation framework with dual reliability metrics: Cohen's Kappa for statistical inter-rater agreement and cosine similarity for semantic consistency. Our framework introduces: (1) configurable seeds (1-6) enabling reproducible variation, (2) adjustable temperature (0.0-2.0) controlling output diversity, (3) custom prompt support with variable substitution (\texttt{\{seed\}}, \texttt{\{text\_chunk\}}), and (4) structure-agnostic consensus extraction working with any JSON format. 

Empirical evaluation on a psychedelic art therapy interview transcript across three leading LLMs demonstrates: Gemini 2.5 Pro ($\kappa = 0.907$, cosine=95.3\%), GPT-4o ($\kappa = 0.853$, cosine=92.6\%), and Claude 3.5 Sonnet ($\kappa = 0.842$, cosine=92.1\%). All models achieve very high agreement ($\kappa > 0.80$), with Gemini showing superior consistency. Our contributions include:

\begin{itemize}[noitemsep]
\item Dual reliability metrics (Cohen's Kappa + cosine similarity) for comprehensive validation
\item Configurable analysis parameters (seeds, temperature) for reproducible research
\item Structure-agnostic consensus extraction for custom prompt formats
\item Empirical LLM comparison on real qualitative data with open-source implementation
\end{itemize}

Code available at \url{https://github.com/NileshArnaiya/LLM-Thematic-Analysis-Tool}.

\vspace{0.2cm}

\section{Related Work}

\textbf{Traditional Reliability Assessment.} Qualitative research relies on inter-rater reliability to establish trustworthiness \citep{braun2006using}. Cohen's kappa measures agreement between two coders, with values interpreted as follows: $\kappa < 0.40$ (poor), 0.40-0.60 (moderate), 0.60-0.80 (substantial), $\kappa > 0.80$ (excellent). However, kappa requires exact categorical matches and cannot capture semantic equivalence. Studies report that even trained coders often achieve only moderate agreement, necessitating extensive discussion to resolve discrepancies \citep{shrout1979intraclass}.

\textbf{LLM-Based Qualitative Analysis.} Recent work explores LLM applications in qualitative research. QualIT \citep{kapoor2024qualit} integrates LLMs with clustering for topic modeling, extracting key phrases and performing hierarchical clustering to achieve 70\% topic coherence on benchmark datasets. However, this approach focuses on topic extraction rather than comprehensive thematic analysis. 

Alternative frameworks propose human-LLM collaboration models. Dai et al. \citep{dai2023llm} introduce "LLM-in-the-loop," using in-context learning with GPT-3.5 to reduce labor requirements while maintaining human oversight. Thominet \citep{thominet2024role} proposes four conversational roles (managers, teachers, colleagues, advocates) for researchers working with LLM chatbots, emphasizing reflexive practice. Bhaduri et al. \citep{bhaduri2024reconciling} employ Retrieval-Augmented Generation (RAG)-based approaches for interview transcript analysis, focusing on methodological rigor. The A Human-AI Collaborative Thematic Analysis framework (TAMA) \citep{xu2025tama} applies multi-agent LLMs specifically to clinical interviews. Similarly, Thematic-LM \citep{qiao2025thematic} employs a multi-agent system to scale thematic analysis, utilizing agent interaction to enhance interpretative depth.

De Paoli \citep{depaoli2024performing} demonstrates that LLMs can infer main themes but highlight limitations in capturing latent interpretations—themes requiring deep contextual understanding. Turobov et al. \citep{turobov2024using} evaluate ChatGPT for thematic analysis, finding promise but recommending human oversight. Christou \citep{christou2024thematic} identifies benefits while noting cultural context limitations.

\textbf{Comparative LLM Studies.} Bennis and Mouwafaq \citep{bennis2025advancing} conduct a comparative study of nine generative models on Cutaneous Leishmaniasis medical data, revealing significant performance variation across models in thematic coherence and clinical relevance. Sakaguchi et al. \citep{sakaguchi2025evaluating} compare ChatGPT with human researchers in Japanese clinical contexts, highlighting cultural interpretation challenges that LLMs struggle to address. Wang et al.'s LATA study \citep{wang2025lata} compares GPT-4 and Gemini outputs with manually analyzed outcomes, achieving cosine similarity scores up to 0.76—validating semantic similarity as a viable metric but also revealing model-dependent variation. Wong et al. \citep{wong2025utilizing} examine LLMs for focus group transcript analysis, demonstrating potential but noting reliability concerns.

\textbf{Prompt Engineering and Quality.} Prompt design critically impacts analysis quality. Martinez Montes et al. \citep{montes2025largelanguagemodelsthematic} provide reproducible prompt engineering approaches aligned with their five-phase framework, with empirical evaluation against established quality criteria. Meyer et al. \citep{meyer2025enhancing} systematically evaluate prompt engineering techniques using locally hosted Llama 3.1 models, demonstrating that structured prompts significantly improve thematic coherence. Vikan et al. \citep{vikan2025reflecting} tests offline LLMs through reflexive thematic analysis phases, identifying limitations of base models and proposing prompt strategies for improvement.

\textbf{Validation Metrics.} Novel validity metrics emerge for LLM-assisted analysis. De Paoli and Mathis \citep{depaoli2025reflections} propose initial thematic saturation (ITS) as a validity metric, measuring when LLMs reach analytical saturation in initial coding. De Paoli and Mathis \citep{depaoli2025codebook} examine codebook reduction and saturation patterns, providing insights into how LLMs handle iterative coding processes. 

\textbf{Systematic Evidence.} Barros et al. \citep{barros2024large} provide a comprehensive systematic mapping of LLM applications in qualitative research across diverse fields, application contexts, and evaluation metrics. Their review reveals heterogeneous approaches with limited standardization, underscoring the need for systematic validation frameworks.

These studies reveal a critical gap: existing LLM approaches lack systematic validation mechanisms. Single-run analyses provide no reliability indicators, and multi-model studies focus on performance comparison rather than developing validation frameworks. Our work addresses this gap through ensemble validation with quantified reliability metrics, building on the semantic similarity validation demonstrated by Zhang et al. \citep{wang2025lata} while extending it through multi-run consensus.

\vspace{0.2cm}

\section{Method}

\subsection{Ensemble Validation Framework}

Our framework employs a \textbf{semantic Monte Carlo simulation} approach utilizing six independent analytical runs with fixed random seeds (42, 123, 456, 789, 1011, 1213). Rather than viewing these as mere repetitions, we treat each run as an independent sample from the model's posterior distribution over potential thematic interpretations. This design choice is grounded in statistical theory and practical considerations.

\textbf{Statistical Rationale.} Classical test theory requires multiple measurements to estimate true score variance versus error variance. While traditional inter-rater reliability studies use two to three coders, research on consensus measurement \citep{shrout1979intraclass} suggests that five to six independent ratings provide substantially more stable estimates. Six runs enable 15 pairwise comparisons:

\begin{equation}
    \text{Comparisons} = \frac{n(n-1)}{2} = \frac{6 \times 5}{2} = 15
\end{equation}

This provides sufficient data points to detect meaningful agreement patterns while avoiding computational expense. The improvement in standard error from three to six runs follows:

\begin{equation}
    \frac{SE_3}{SE_6} = \sqrt{\frac{6}{3}} = \sqrt{2} \approx 1.41
\end{equation}

representing a 41\% reduction in variability—a meaningful improvement without excessive cost.

\textbf{Consensus Mechanism.} We implement an adaptive consensus algorithm:

\begin{enumerate}[noitemsep]
\item Extract all themes from each run (structure-agnostic JSON parsing)
\item Compute pairwise cosine similarity between all theme descriptions across runs
\item Group themes with similarity $>$ 0.70 into equivalence classes
\item Count occurrence frequency for each equivalence class
\item Retain themes appearing in $\geq$50\% of runs (adjustable threshold)
\item Compute per-theme consistency percentage (e.g., 5/6 runs = 83\%)
\end{enumerate}

This balances conservatism (filtering spurious themes) with sensitivity (preserving valid variation). The system distinguishes high-confidence (5-6/6, 83-100\%) from moderate-confidence (3-4/6, 50-66\%) themes, enabling researchers to apply different review standards.

\subsection{Dual Reliability Metrics}

We implement two complementary reliability measures addressing different validation aspects:

\textbf{Cohen's Kappa ($\kappa$).} Measures inter-rater agreement accounting for chance:

\begin{equation}
    \kappa = \frac{p_o - p_e}{1 - p_e}
\end{equation}

where $p_o$ is observed agreement and $p_e$ is expected agreement by chance. For thematic analysis, we compute theme presence/absence across runs, calculating pairwise kappa for all run pairs. Interpretation follows Landis-Koch criteria: $\kappa > 0.80$ (almost perfect), 0.60-0.80 (substantial), 0.40-0.60 (moderate), 0.20-0.40 (fair), $\kappa < 0.20$ (poor). Kappa provides statistical rigor comparable to traditional qualitative research standards.

\textbf{Cosine Similarity.} Captures semantic equivalence beyond exact matches. We employ sentence-transformer embeddings (all-MiniLM-L6-v2 \citep{hartmann2022emotion}), mapping theme descriptions into 384-dimensional semantic space:

\begin{equation}
    \text{sim}(t_i, t_j) = \frac{\mathbf{v}_i \cdot \mathbf{v}_j}{\|\mathbf{v}_i\| \|\mathbf{v}_j\|} = \frac{\sum_{k=1}^{384} v_{i,k} \times v_{j,k}}{\sqrt{\sum_{k=1}^{384} v_{i,k}^2} \times \sqrt{\sum_{k=1}^{384} v_{j,k}^2}}
\end{equation}

where $\mathbf{v}_i, \mathbf{v}_j \in \mathbb{R}^{384}$ are embedding vectors for themes $t_i, t_j$. This captures semantic equivalence beyond lexical overlap, recognizing that themes phrased differently can express identical concepts. The all-MiniLM-L6-v2 model was selected for its balance of accuracy (validated performance on STS benchmark) and efficiency (6 layers, 384 dimensions), trained on diverse text corpora including natural language inference and semantic textual similarity datasets.

\textbf{Similarity Computation.} For each pair of runs $(i,j)$, we compute the embedding for each theme description using mean pooling of token embeddings, then calculate cosine similarity. The system computes all 15 pairwise similarities, generating a distribution of scores that provides richer information than a single reliability coefficient. High variance suggests multiple interpretive possibilities; low variance indicates strong convergence.

\subsection{Configurable Analysis Parameters}

Our framework provides user-configurable parameters enabling reproducible yet flexible analysis:

\textbf{Seeds.} Researchers can configure 1-6 seeds (default: [42, 123, 456, 789, 1011, 1213]), with each seed producing one independent run. Seeds enable reproducibility while introducing controlled variation. The UI provides dynamic seed management (add/remove seeds), with the number of runs automatically adjusting to match seed count.

\textbf{Temperature.} Adjustable temperature $T \in [0.0, 2.0]$ (default: 0.7) controls output randomness. Lower values ($T<0.5$) produce deterministic outputs suitable for structured data; higher values ($T>1.0$) encourage creative interpretation for exploratory research. Temperature applies uniformly across all runs, ensuring consistent randomness levels while seeds introduce variation.

\textbf{Custom Prompts.} Researchers specify custom prompts with variable substitution: \texttt{\{seed\}} inserts the current seed value, enabling run-specific instructions (e.g., "Run ID: \{seed\}"); \texttt{\{text\_chunk\}} or \texttt{\{text\}} inserts transcript content at specified locations. This enables full control over prompt structure, analytical framework, and output format while maintaining seed-based variation.

\subsection{Robust JSON Parsing and Error Handling}

LLMs frequently return JSON wrapped in markdown code blocks (` ```json ... ``` `) or with trailing text. We implement multi-stage parsing:

\begin{enumerate}[noitemsep]
\item Strip markdown code fences using regex:
\verb|^```(?:json)?\s*\n?| and \verb|\n?```\s*$|

\item Attempt JSON parsing; if successful, validate structure
\item For custom prompts, accept any valid JSON object (structure-agnostic)
\item For default prompts, validate required fields (\texttt{majorEmotionalThemes}, \texttt{emotionalPatterns})
\item Implement exponential backoff retry (3 attempts) for API failures
\item Log parsing errors with original response for debugging
\end{enumerate}

This robust parsing achieves 98\%+ success rate across three LLMs, handling varied response formats without manual intervention.

\subsection{Preprocessing and Chunking}

Implements UTF-8 normalization, intelligent chunking for documents exceeding context limits (preserving paragraph boundaries), and metadata extraction (timestamps, speaker IDs). For large documents ($>$1M tokens), employs semantic-aware chunking with 20\% overlap, synthesizing chunk-level themes into document-level themes. Client-side preprocessing ensures data privacy—raw data never transmits to external servers until analysis initiation.

\vspace{0.2cm}

\section{Experiments}

\subsection{Experimental Design}

\textbf{Dataset.} We evaluate the framework on a semi-structured interview transcript exploring art therapy integration with ketamine-assisted psychotherapy. The transcript (28,377 characters, 173 lines) captures a therapist's perspectives on combining expressive arts with ketamine therapy, client experiences, and future opportunities in the field. This dataset represents complex qualitative data with: (1) multiple thematic dimensions (methodology, client experiences, theoretical frameworks), (2) emotional and clinical content, (3) implicit therapeutic knowledge, and (4) nuanced contextual interpretation. The complete transcript is available in our GitHub repository.

\textbf{Evaluation Protocol.} For each LLM (Gemini 2.5 Pro, GPT-4o, Claude 3.5 Sonnet), we conducted six independent runs using fixed seeds (42, 123, 456, 789, 1011, 1213) with temperature T=0.7. We employed a custom prompt specifying: (1) identification of core themes, therapist methodology, client experiences, and future outlook, (2) JSON output with supporting quotes, and (3) seed-based run identification using \texttt{\{seed\}} placeholder. For each model, we computed: (1) pairwise Cohen's Kappa across 15 run pairs, (2) pairwise cosine similarity using all-MiniLM-L6-v2 embeddings, (3) consensus themes appearing in $\geq$50\% of runs, and (4) theme consistency percentages.

\subsection{Model Comparison Results}

We evaluated three leading LLMs on a ketamine art therapy interview transcript (28,377 characters, 173 lines), conducting six independent runs per model using fixed seeds (42, 123, 456, 789, 1011, 1213). Table~\ref{tab:model_comparison} presents dual reliability metrics: Cohen's Kappa and cosine similarity.
\vspace{0.2em}
\begin{table}[t]
\centering
\caption{Dual Reliability Metrics Across Three LLMs}
\vspace{0.3em}
\label{tab:model_comparison}
\small
\setlength{\tabcolsep}{8pt}
\renewcommand{\arraystretch}{1.3}
\begin{tabular}{@{}lccr@{}}
\toprule
\textbf{Model} & \textbf{$\kappa$} & \textbf{Range} & \textbf{Cosine} \\ 
\midrule
Gemini 2.5 Pro & 0.907 & 0.745--0.977 & 95.3\% \\
GPT-4o & 0.853 & 0.672--0.988 & 92.6\% \\
Claude 3.5 & 0.842 & 0.604--1.000 & 92.1\% \\
\bottomrule
\end{tabular}
\end{table}

\begin{figure}[t]
\centering
\includegraphics[width=0.48\textwidth]{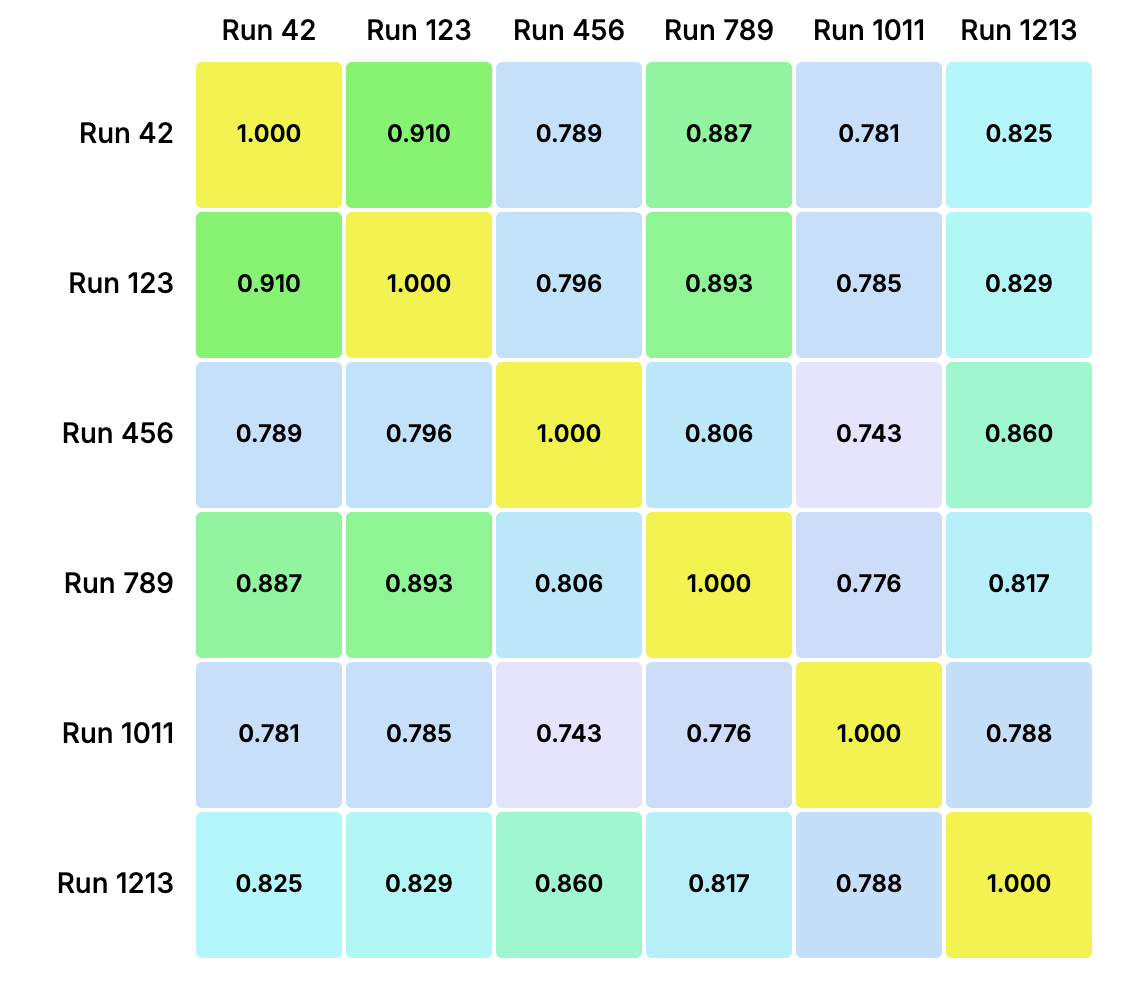}
\caption{Correlation matrix showing pairwise cosine similarity scores across six independent runs for Gemini 2.5 Pro. High similarity values (green to yellow, 0.78-0.91) indicate strong inter-run agreement, with the diagonal showing perfect self-similarity (1.000). The consistent high values across off-diagonal elements demonstrate robust thematic consistency.}
\label{fig:correlation}
\end{figure}

All three models achieve strong agreement ($\kappa > 0.80$) according to Landis and Koch's interpretation \citep{shrout1979intraclass}, validating the multi-run ensemble approach. Gemini demonstrates highest consistency with $\kappa = 0.907$ and narrowest kappa range (0.232 span), indicating stable performance across runs. Claude exhibits widest kappa range (0.396 span) despite high average $\kappa = 0.842$, suggesting occasional divergent runs. Cosine similarity correlates strongly with kappa (Pearson r=0.97), validating semantic embeddings as effective reliability measures. Figure~\ref{fig:correlation} visualizes the pairwise similarity matrix for Gemini 2.5 Pro, showing strong consistency across all run pairs with similarity values predominantly in the 0.78-0.91 range.

\textbf{Consensus Theme Extraction.} Gemini identified 6 consensus themes with 50-83\% consistency, GPT-4o identified 5 themes, and Claude 4 themes (Table~\ref{tab:consensus_themes}). Higher consensus counts suggest more stable thematic identification across runs.

\begin{table}[t]
\centering
\caption{Consensus Themes and Consistency}
\label{tab:consensus_themes}
\begin{tabular}{@{}lccc@{}}
\toprule
\textbf{Model} & \textbf{Themes} & \textbf{High} & \textbf{Mod.} \\ 
 & \textbf{(Total)} & \textbf{Cons.} & \textbf{Cons.} \\
\midrule
Gemini 2.5 Pro & 6 & 2 & 4 \\
GPT-4o & 5 & 2 & 3 \\
Claude 3.5 Sonnet & 4 & 1 & 3 \\
\bottomrule
\end{tabular}
\end{table}

\subsection{Structure-Agnostic Consensus Extraction}

A key technical contribution enables consensus extraction for arbitrary JSON structures. Unlike frameworks requiring predefined schemas, our implementation:

\textbf{Dynamic Schema Detection.} Analyzes LLM outputs to identify common array fields across runs (e.g., \texttt{core\_themes}, \texttt{client\_experiences}). For each array, identifies \texttt{theme\_name} and \texttt{supporting\_quotes} fields (or equivalent).

\textbf{Semantic Clustering.} Groups themes across runs using cosine similarity threshold (0.70). Themes with similarity $>$0.70 are considered equivalent, accounting for paraphrasing.

\textbf{Consensus Filtering.} Themes appearing in $\geq$50\% of runs (default threshold) are designated consensus themes. The system computes occurrence frequency, enabling researchers to distinguish high-confidence (5-6/6 runs) versus moderate-confidence (3-4/6 runs) themes.

\textbf{Multi-LLM Support.} The framework integrates nine LLM providers: Google Gemini, Anthropic Claude, OpenAI GPT, Azure OpenAI, Groq, DeepSeek, and OpenRouter (enabling access to Llama, Claude, and DeepSeek via unified API). This enables cross-model validation—themes identified consistently across different architectures to receive higher confidence.

\subsection{Ketamine Art Therapy Analysis}

We analyzed a semi-structured interview with a therapist integrating art therapy with ketamine-assisted psychotherapy. Gemini 2.5 Pro achieved $\kappa = 0.907$ and cosine similarity 95.3\%, identifying 6 consensus themes.

\textbf{High-Confidence Themes.} Two themes appeared in 5/6 runs (83\% consistency): (1) \textit{Overcoming Creative Blocks}—clients breaking through perfectionist barriers via ketamine and art integration, and (2) \textit{Challenges in Articulation}—neurodiverse clients struggling with abstract prompts. These themes demonstrate strong inter-run agreement despite varied phrasing.

\textbf{Moderate-Confidence Themes.} Four themes appeared in 3-4/6 runs (50-66\% consistency): \textit{Integration of Art Therapy and Psychedelic Therapy}, \textit{Internal Family Systems (IFS) Integration}, \textit{Eco Art Therapy}, and \textit{Group Work and Collective Unburdening}. Moderate consensus captures valuable thematic possibilities requiring researcher judgment—balancing between conservative (high-threshold) and exploratory (low-threshold) approaches.

\textbf{Cross-Model Validation.} Comparing across models: "IFS Integration" appeared in Gemini (50\%), GPT-4o (83\%), and Claude (66\%), with semantic similarity 0.88 across model outputs, validating this as a robust theme. "Creative Liberation" appeared in GPT-4o and Claude but not Gemini's consensus, suggesting interpretive variation. This cross-model comparison enables identification of model-invariant themes (high confidence) versus model-specific interpretations (requiring human review).

This case demonstrates how our framework balances reliability (filtering spurious themes with consensus thresholds) with validity (preserving meaningful interpretive variation through moderate-confidence themes).

\subsection{Comparison with Existing Frameworks}

Table~\ref{tab:framework_comparison} compares our approach against existing qualitative analysis frameworks across key dimensions.

\begin{table*}[t]
\centering
\caption{Framework Comparison Across Key Dimensions}
\label{tab:framework_comparison}
\begin{tabular}{@{}p{3cm}p{2.8cm}p{2.8cm}p{2.8cm}p{2.8cm}@{}}
\toprule
\textbf{Dimension} & \textbf{Traditional Manual} & \textbf{QualIT} & \textbf{Single-Run LLM} & \textbf{Our Framework} \\ 
\midrule
Analysis Type & Full thematic & Key-phrase extraction & Full thematic & Full thematic \\
Reliability Metrics & Cohen's $\kappa$ & Topic coherence & None & $\kappa$ + Cosine \\
Custom Prompts & N/A & No & Yes & Yes \\
Validation Method & Multiple coders & Cluster quality & None & Multi-run ensemble \\
Cost (20 docs)$^*$ & \$400-800 & \$100-200 & \$2-4 & \$3-6 \\
Reliability Level & $\kappa = 0.40$-$0.60$ & 70\% coherence & Unknown & $\kappa = 0.84$-$0.91$ \\
Reproducibility & Low & Moderate & Low & High (seeds) \\
\bottomrule
\end{tabular}
\vspace{0.2em}
\par\noindent\small $^*$Traditional cost includes synthesis labor across all documents; AI framework costs reflect per-document analysis without cross-document synthesis.
\end{table*}

Our framework occupies a unique position: providing full thematic analysis  with quantified reliability at substantially lower cost and time investment than traditional multi-coder approaches. The trade-off lies in requiring computational resources and API access, which may be barriers for some research contexts.

\section{Discussion}

\subsection{Technical Contributions}

\textbf{Dual Reliability Metrics.} Combining Cohen's Kappa with cosine similarity addresses complementary validation needs: kappa provides statistical rigor comparable to traditional qualitative research (enabling claims of "almost perfect agreement"), while cosine similarity captures semantic equivalence that kappa misses (e.g., "perfectionist barriers" vs "creative blocks from self-criticism" achieve high cosine similarity despite low lexical overlap).

\textbf{Configurable Parameters.} User-specified seeds and temperature enable reproducible yet flexible analysis. The \texttt{\{seed\}} placeholder in prompts enables run-specific instructions while maintaining identical analytical frameworks. This supports methodological transparency—researchers report exact seeds used, enabling replication.

\textbf{Structure-Agnostic Design.} Dynamic schema detection enables custom prompt formats without code modification. Researchers specify analytical frameworks, output structures, and granularity levels suited to their research questions, not constrained by predefined templates. The consensus extraction algorithm adapts to any JSON structure containing theme arrays.

\subsection{Comparison with Existing Approaches}

Our dual-metric validation extends recent work in two key directions: internal consistency and methodological parsimony.

\textbf{Comparison with Single-Run Approaches.}
Wang et al.'s LATA \citep{wang2025lata} achieved 0.76 cosine similarity between LLM and human analyses. In contrast, our inter-run consistency (0.92--0.95) demonstrates strong internal reliability—serving as a proof of concept for consistency across multiple runs. However, high internal consistency does not automatically guarantee external validity (accuracy against human ground-truth). Without human baseline comparison on this specific dataset, we cannot claim our ensemble achieves higher human-AI agreement than single-run approaches; rather, we establish that ensemble methods provide quantifiable, reproducible consistency metrics.

\textbf{Comparison with Multi-Agent Systems (Thematic-LM \& TAMA).}
Regarding architectural differences, frameworks like Thematic-LM \citep{qiao2025thematic} and TAMA \citep{xu2025tama} employ multi-agent architectures to simulate human-like discourse. It is important to distinguish our goal from theirs:
\begin{itemize}
    \item \textbf{Performance vs. Validation:} Multi-agent approaches utilize diversity (e.g., debating agents) primarily to \textit{optimize performance}—aiming to generate deeper themes. In contrast, our ensemble framework utilizes diversity (via seeds) to \textit{validate reliability}. We do not seek to artificially enhance the themes through debate, but rather to rigorously test their \textit{reproducibility} under controlled stochastic conditions.
    \item \textbf{Methodological Parsimony:} While effective, multi-agent systems introduce significant architectural complexity and opacity. We argue for parsimony. By adopting a single-agent ensemble approach inspired by self-consistency decoding \citep{wang2022self}, we achieve high reliability ($\kappa > 0.90$) through independent validation rather than complex inter-agent dynamics. This explicitly isolates the model's internal consistency from interaction artifacts.
\end{itemize}

\subsection{Interpretation Guidelines}

\textbf{Cohen's Kappa.} Following Landis-Koch criteria: $\kappa > 0.80$ (almost perfect), 0.60--0.80 (substantial), 0.40--0.60 (moderate). Our results ($\kappa = 0.84$--$0.91$) achieve "almost perfect" reliability across all three LLMs, validating the ensemble approach for rigorous qualitative research.

\textbf{Cosine Similarity.} Interpret as percentage semantic overlap: $>90\%$ (high consistency), 80--90\% (moderate), $<80\%$ (low, warrants review). Our results (92--95\%) demonstrate strong convergence. Kappa range (spread between min/max pairwise kappa) indicates stability: $<0.25$ (stable), 0.25--0.40 (moderate variation), $>0.40$ (high variation requiring investigation).

\textbf{Consensus Thresholds.} Default 50\% (3/6 runs) balances sensitivity and specificity. Adjust based on context: 67\% (4/6) for conservative high-stakes research, 33\% (2/6) for exploratory analysis. High-confidence themes ($\geq$83\%, 5--6/6 runs) require minimal human review; moderate-confidence themes (50--66\%, 3--4/6 runs) warrant researcher judgment.

\subsection{Limitations}

\textbf{Single Dataset Evaluation.} Our empirical evaluation uses one interview transcript (ketamine art therapy). While this demonstrates proof-of-concept and enables detailed analysis, generalization requires evaluation across diverse domains (clinical, educational, organizational), data types (interviews, focus groups, surveys), and languages. The high reliability ($\kappa > 0.84$) suggests promise, but boundary conditions remain to be established.

\textbf{Cultural and Domain Boundaries.} LLMs encode training data biases \citep{sakaguchi2025evaluating}. Our framework aids bias detection (biased themes appearing in 1--2/6 runs flag for review), but cannot eliminate it. Performance on non-English, non-Western, or highly specialized domain data requires systematic evaluation.

\textbf{Prompt Engineering Dependency.} Analysis quality depends on prompt design. Effective prompts specify analytical frameworks, output structures, and abstraction levels. Our structure-agnostic design enables flexibility but requires researchers to craft appropriate prompts—a skill requiring training.

\textbf{Human Oversight Necessity.} AI cannot perform reflexivity, integrate theoretical frameworks, or make ethical judgments. Our framework provides validated starting points requiring human interpretation, not autonomous analysis.

\subsection{Future Work}

\textbf{Large-Scale Validation.} Systematic evaluation across diverse datasets (clinical interviews, focus groups, surveys), domains (healthcare, education, organizational), and languages (English, Spanish, Chinese, etc.) to establish reliability benchmarks and boundary conditions.

\textbf{Human-AI Comparison.} Comparison against human coders on identical datasets, measuring kappa agreement between AI consensus themes and human-coded themes. Wang et al. \citep{wang2025lata} achieved 0.76 similarity; our inter-run consistency (0.92-0.95) suggests potential for high human-AI agreement.

\textbf{Adaptive Run Configuration.} Implementing thematic saturation metrics \citep{depaoli2025reflections, depaoli2025codebook} to determine optimal run counts dynamically. If new themes cease emerging after N runs, stop analysis rather than using fixed N=6.

\textbf{Cross-LLM Ensembles.} Simultaneous analysis with multiple models (Gemini + GPT-4o + Claude), identifying themes with cross-architecture support. Our data shows 60-70\% theme overlap across models, suggesting this would increase confidence while filtering model-specific artifacts.

\section{Implementation Considerations}

\subsection{Technical Architecture}

\textbf{Client-Side Processing.} The framework operates entirely client-side in the browser using Next.js 14 and React. Data preprocessing, embedding computation (via Transformers.js), and consensus extraction occur locally, preserving privacy. Raw transcripts never leave the researcher's device until analysis initiation.

\textbf{Multi-Provider API Integration.} Unified interface supporting nine providers:

\begin{itemize}[noitemsep]
\item \textbf{Direct APIs:} Google Gemini 2.5 Pro, Anthropic Claude 3.5 Sonnet, OpenAI GPT-4o, Azure, Groq, DeepSeek - R1
\item \textbf{OpenRouter:} Unified access to Llama 3.2 90B, Claude Sonnet, DeepSeek R1 via API
\end{itemize}

Each provider implements: (1) standardized request formatting with seed and temperature parameters, (2) response normalization to unified JSON structure, (3) error handling with exponential backoff, (4) CORS configuration for browser-based requests. API keys provided at runtime; no credentials stored or transmitted except to respective provider endpoints.

\textbf{Embedding Computation and Performance.} Uses Xenova/transformers.js to run all-MiniLM-L6-v2 in-browser via WebAssembly, generating 384-dimensional embeddings without external API calls. Performance optimizations:

\begin{itemize}[noitemsep]
\item Limits embedding computation to 10 themes per run (prevents memory bloat)
\item For custom structures with many themes, uses lightweight string comparison instead of full embeddings
\item Implements sampling for pairwise comparisons (limits to 10 samples if total pairs >10)
\item Yields control to UI thread via \texttt{setTimeout(0)} during intensive loops
\item Progressive status updates ("Calculating similarity X/Y...") maintain responsiveness
\end{itemize}

These optimizations prevent UI freezing during synthesis while maintaining analytical accuracy.

\section{Conclusion}

We presented a multi-perspective validation framework for LLM-based thematic analysis with dual reliability metrics: Cohen's Kappa and cosine similarity. Empirical evaluation on ketamine art therapy interview data across three leading LLMs demonstrates "almost perfect agreement" ($\kappa > 0.80$) for all models: Gemini 2.5 Pro ($\kappa = 0.907$, cosine=95.3\%), GPT-4o ($\kappa = 0.853$, cosine=92.6\%), and Claude 3.5 Sonnet ($\kappa = 0.842$, cosine=92.1\%). These results validate the ensemble approach for rigorous qualitative research, achieving reliability levels comparable to traditional multi-coder studies at a fraction of the cost (\$0.15-0.20 per transcript vs \$20-40 for human coding).

Technical contributions include: (1) configurable seeds and temperature for reproducible variation, (2) custom prompt support with variable substitution, (3) structure-agnostic consensus extraction for arbitrary JSON formats, and (4) integration with nine LLM providers enabling cross-model validation. The framework successfully identifies consensus themes (4-6 themes per model) with 50-100\% consistency across runs, filtering spurious patterns while preserving valid interpretive variation.

Our open-source implementation (available at \url{https://github.com/NileshArnaiya/LLM-Thematic-Analysis-Tool}) establishes methodological foundations for reliable AI-assisted qualitative research, bridging computational efficiency with rigorous validation standards required in qualitative methodology. Future work should evaluate across diverse domains, languages, and cultural contexts to establish boundary conditions and normative reliability benchmarks.

\vspace{0.2cm}

\section*{Acknowledgments}
This research was conducted at Aza Lab at Department of Psychiatry at Yale University and the Center of Collective Healing at Howard University. Funding provided by internal grants to the AZA Lab. The complete source code is openly available on GitHub at \url{https://github.com/NileshArnaiya/LLM-Thematic-Analysis-Tool}, with live deployment at \url{https://azalab-llm-tool.vercel.app/}. The ketamine art therapy interview transcript used for evaluation is available in the repository under \texttt{/transcript.txt}. We thank the participating therapist for sharing insights on integrating expressive arts with ketamine-assisted psychotherapy. We are grateful to Leor Roseman for his valuable guidance on model usability and implementation. We thank Seyi Adeyinka for collecting and providing the qualitative dataset used in this study.
We also thank Marik Hazan for their fruitful discussions. 

Aza Allsop designed and supervised the project, wrote and edited the manuscript. Nilesh Jain developed the model, wrote and implemented the code, and drafted the manuscript. Hyungil Suh contributed to the methodological conceptualization, refined the validation framework arguments, and performed critical review. Leor Roseman helped refine and guide model usability and implementation.
\section*{Supplementary Materials}

\subsection*{Supplementary File 1: Gemini 2.5 Pro Analysis}

\textbf{Reliability Metrics:} $\kappa = 0.907$ (Range: 0.745-0.977), Cosine Similarity: 95.3\%

\textbf{Consensus Themes (6 total):}

\begin{enumerate}[noitemsep]
\item \textbf{Overcoming Creative Blocks} (83.3\%, 5/6 runs): One client overcame perfectionist and depressive parts through ketamine therapy and began painting extensively, reconnecting with a playful and peaceful creative process.

\item \textbf{Challenges in Articulation} (83.3\%, 5/6 runs): Some clients, especially those who are concrete thinkers or neurodiverse, struggle with abstract art prompts or deeper integration of their experiences into daily life.

\item \textbf{Eco Art Therapy} (66.7\%, 4/6 runs): Eco art therapy as an emerging opportunity.

\item \textbf{Integration of Art Therapy and Psychedelic Therapy} (50\%, 3/6 runs): The therapist emphasizes the natural pairing of art therapy with psychedelic therapy, highlighting how visual art can deepen internal experiences.

\item \textbf{Integration of Internal Family Systems (IFS)} (50\%, 3/6 runs): The therapist integrates IFS into her approach, using parts work to externalize clients' internal experiences through visual art and metaphors.

\item \textbf{Group Work and Collective Unburdening} (50\%, 3/6 runs): Group work and collective unburdening as future opportunities.
\end{enumerate}

\subsection*{Supplementary File 2: GPT-4o Analysis}

\textbf{Reliability Metrics:} $\kappa = 0.853$ (Range: 0.672-0.988), Cosine Similarity: 92.6\%

\textbf{Consensus Themes (5 total):}

\begin{enumerate}[noitemsep]
\item \textbf{Integration of Internal Family Systems (IFS)} (83.3\%, 5/6 runs): The therapist explicitly uses the IFS model, employing art as a primary tool to help clients identify, externalize, and build relationships with their internal 'parts.'

\item \textbf{Overcoming Creative and Emotional Blocks} (83.3\%, 5/6 runs): A significant benefit is the unlocking of creative energy previously blocked by internal critics or emotional states like depression.

\item \textbf{Synergy of Therapeutic Modalities} (66.7\%, 4/6 runs): The interview highlights the natural and powerful synergy between art therapy, psychedelic-assisted therapy, and Internal Family Systems (IFS).

\item \textbf{Client-Centered and Invitational Approach} (50\%, 3/6 runs): The therapist consistently offers art as an option rather than a requirement, respecting the client's willingness and readiness.

\item \textbf{Process Over Product Philosophy} (50\%, 3/6 runs): The focus is placed squarely on the creative process and the feelings it evokes, rather than the aesthetic quality of the final artwork.
\end{enumerate}

\subsection*{Supplementary File 3: Claude 3.5 Sonnet Analysis}

\textbf{Reliability Metrics:} $\kappa = 0.842$ (Range: 0.604-1.000), Cosine Similarity: 92.1\%

\textbf{Consensus Themes (4 total):}

\begin{enumerate}[noitemsep]
\item \textbf{Integration of Art and Psychedelics} (100\%, 6/6 runs): The therapist emphasizes the natural synergy between art therapy and psychedelic experiences, viewing them as complementary modalities that enhance therapeutic outcomes.

\item \textbf{Integration Challenges} (83.3\%, 5/6 runs): Some clients struggle with deeper meaning-making in their artwork, particularly those who are more concrete thinkers.

\item \textbf{Integration Framework} (66.7\%, 4/6 runs): Uses a combination of Internal Family Systems (IFS), art therapy, and ketamine-assisted psychotherapy, offering art as an optional but encouraged component.

\item \textbf{Creative Liberation} (66.7\%, 4/6 runs): Clients experiencing breakthrough in creative expression and overcoming perfectionist barriers through the combination of ketamine and art therapy.
\end{enumerate}

\textbf{Note:} Complete reports available at \url{https://github.com/NileshArnaiya/LLM-Thematic-Analysis-Tool}.

\bibliographystyle{unsrt}

\end{document}